\documentclass{bmvc2k}

\usepackage{amsmath}
\usepackage{algorithmic}
\usepackage{amssymb}
\usepackage{booktabs}
\usepackage{multirow}
\usepackage{colortbl}
\usepackage{diagbox}
\usepackage{array}
\usepackage{floatrow}
\newfloatcommand{capbtabbox}{table}[][\FBwidth]

\newcommand{\doubleQuote}[1]{\lq\lq{#1}\rq\rq}

\title{Light Cascaded Convolutional Neural Networks for Accurate Player Detection}
\addauthor{Keyu Lu}{keyu.lu@nudt.edu.cn}{1}
\addauthor{Jianhui Chen}{jhchen14@cs.ubc.ca}{2}
\addauthor{James J. Little}{little@cs.ubc.ca}{2}
\addauthor{Hangen He}{hangenhe@nudt.edu.cn}{1}

\addinstitution{
 College of Mechatronic Engineering and Automation\\
 National University of Defense Technology\\
 Changsha, China
}

\addinstitution{
 Department of Computer Science\\
 University of British Columbia\\
 Vancouver, Canada
}

\runninghead{K. Lu, J. Chen, J. J. Little, H. He}{Light Cascaded CNNS for Player Detection}
\def\etal{\emph{et al}\bmvaOneDot}

\makeatletter
\g@addto@macro\normalsize{%
	\setlength\abovedisplayskip{3pt}
	\setlength\belowdisplayskip{3pt}
	\setlength\abovedisplayshortskip{3pt}
	\setlength\belowdisplayshortskip{3pt}
}
\makeatother

\begin{document}

\maketitle

\footnotetext[1]{{This work was performed while Keyu Lu was visiting the LCI lab at UBC.}}

% has particular challenges from highly dynamic group sports such as soccer and basketball.
\begin{abstract}
	
Vision based player detection is important in sports applications. Accuracy, efficiency, and low memory consumption are desirable for real-time tasks such as intelligent broadcasting and automatic event classification. In this paper, we present a cascaded convolutional neural network (CNN) that satisfies all three of these requirements. Our method first trains a binary (player/non-player) classification network from labeled image patches. Then, our method efficiently applies the network to a whole image in testing. We conducted experiments on basketball and soccer games. Experimental results demonstrate that our method can accurately detect players under challenging conditions such as varying illumination, highly dynamic camera movements and motion blur. Comparing with conventional CNNs, our approach achieves state-of-the-art accuracy on both games with $1000 \times$ fewer parameters (i.e., it is \textit{light}).

\end{abstract}

%-------------------------------------------------------------------------

%\vspace{-2mm}
\section{Introduction}
\label{sec:intro}
%\vspace{-2mm}
Player detection from images and videos is essential for a number of applications. For example, intelligent broadcast systems use player locations to guide the viewpoints of broadcasting cameras~\cite{chen2016where}. Furthermore, player detection provides metadata for player tracking, player pose estimation and team strategy analysis \cite{thomas2017computer}. Player detection, as a subcategory of people detection, has been extensively studied. For example, background subtraction based methods~\cite{carr2012monocular,Parisot2017} have been applied to basketball games and achieved real-time response. However, these methods assume the camera is static or moves slowly so that they can robustly detect foreground objects. Some learning-based methods such as Faster R-CNN~\cite{FasterRCNN16} and YOLO~\cite{YOLO15} can be also adapted to detect players with high detection accuracy but they may miss distant players because of low pixel resolution of these players. 

Our work is inspired by CNN-based object detection and cascaded learning. Recently, a number of CNN-based approaches have achieved excellent performance for general object detection~\cite{Fast_RCNN15,FasterRCNN16,YOLO15,SSD15,Anelia15,FanYang16,Hongwei16} or pedestrian detection~\cite{liu2016multispectral,zhang2016isfaster}. However, to the best of our knowledge, there are few deep neural networks specifically designed for player detection.  

\begin{figure}[t]
	\centering
	\subfigure[Varying appearance]{\includegraphics[width=0.235\textwidth,keepaspectratio]{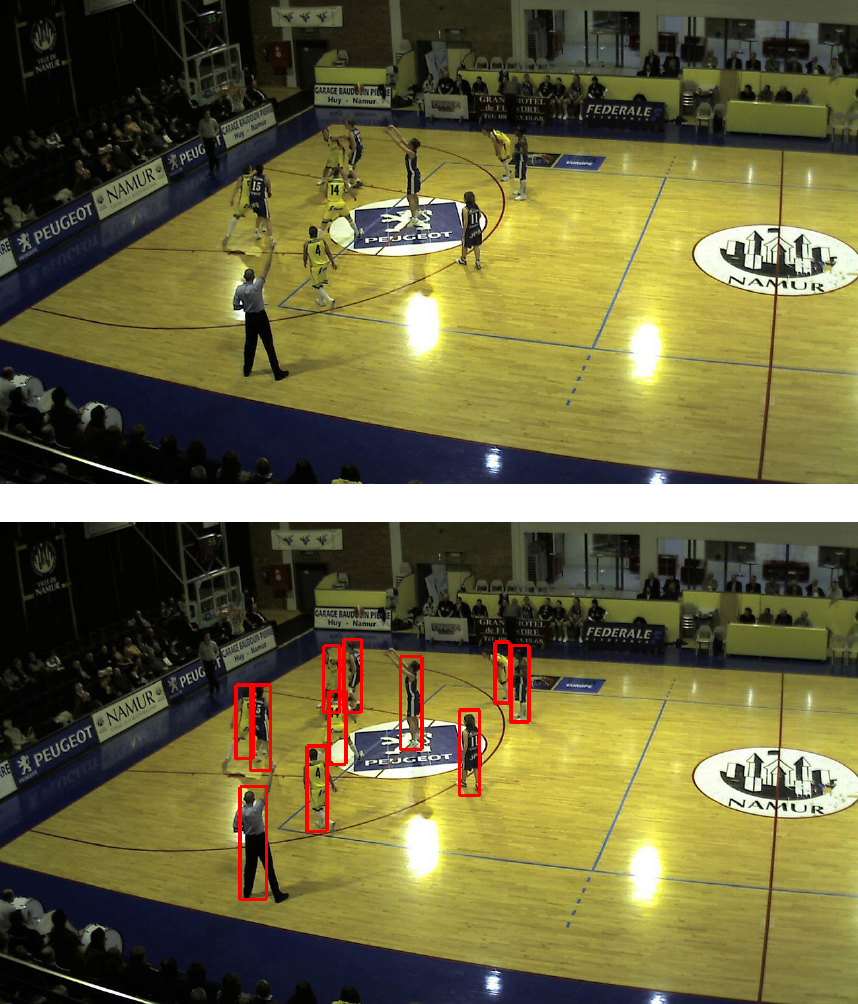}
		\label{fig_hard_1}}
	\subfigure[Cluttered background]{\includegraphics[width=0.235\textwidth,keepaspectratio]{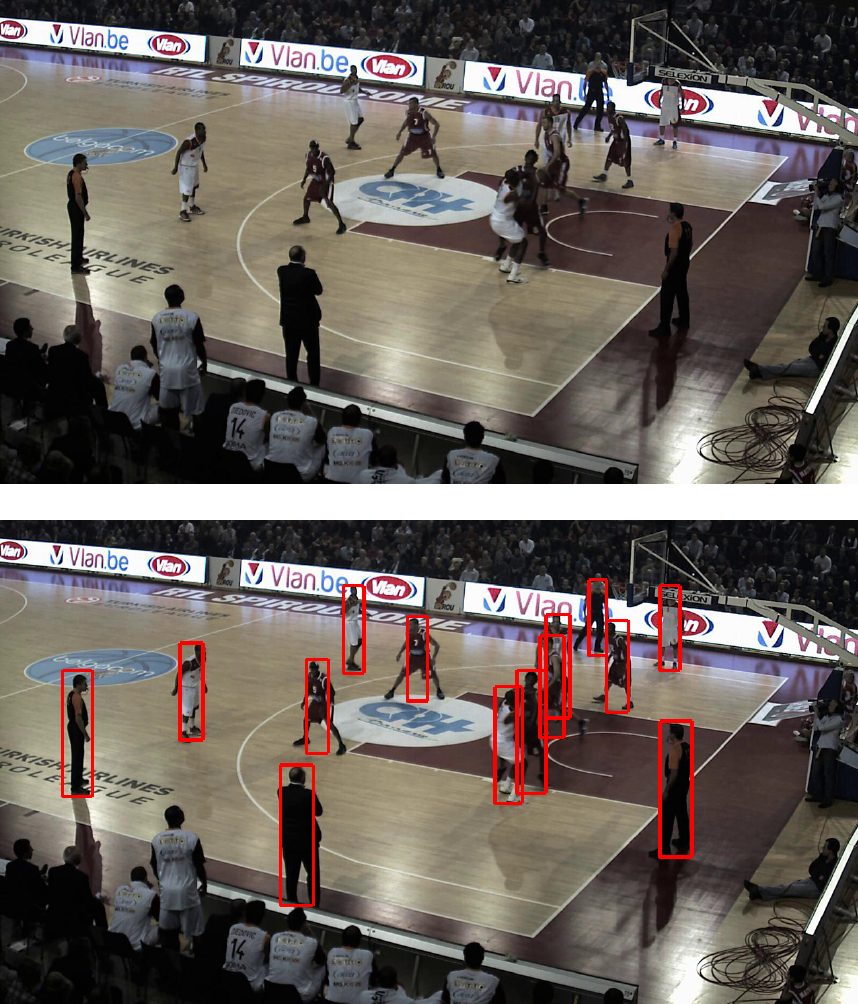}
		\label{fig_hard_2}}
	\subfigure[Distant players]{\includegraphics[width=0.235\textwidth,keepaspectratio]{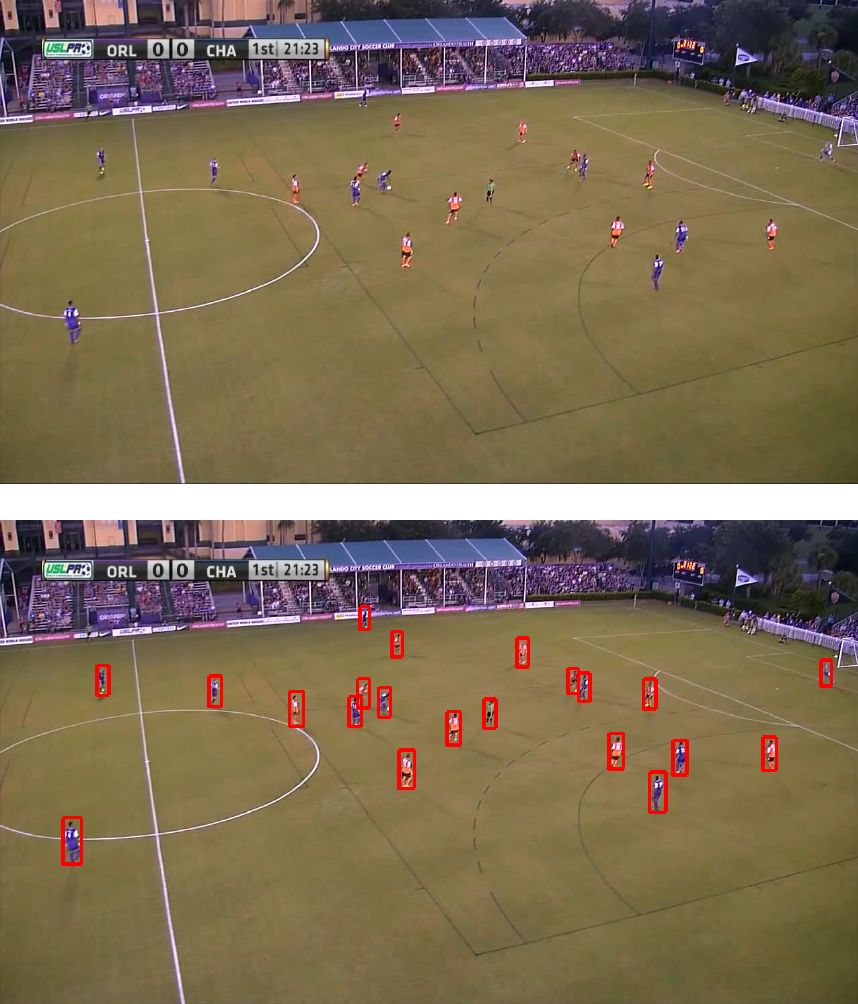}
		\label{fig_hard_5}}
	\subfigure[Motion blur]{\includegraphics[width=0.235\textwidth,keepaspectratio]{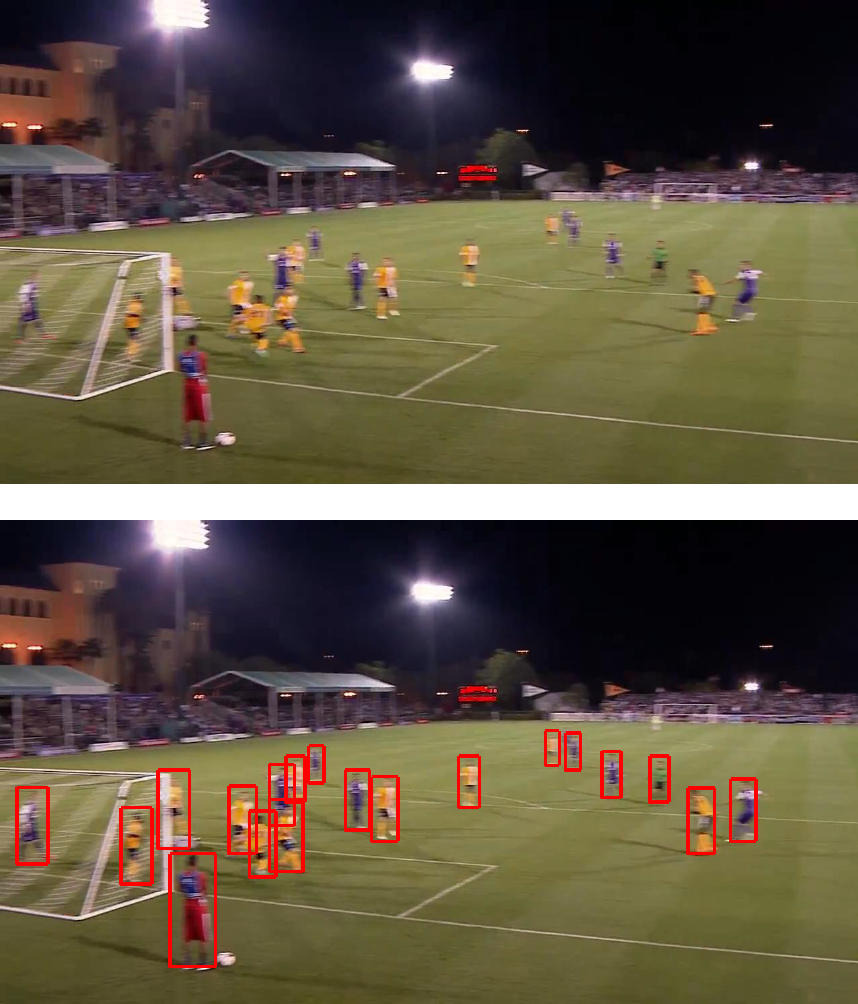}
		\label{fig_hard_6}}
	\vspace{-2.6ex}
	\caption{\textbf{Examples of player detection challenges.} The second row shows the detection results of our approach.}
	\label{fig:challenge}
\end{figure}

Compared with pedestrian detection, player detection is more challenging in terms of body and camera motions~\cite{Manafifard17}. For example, basketball players quickly jump, extend their hands and twist their bodies (see Figure \ref{fig:challenge}). Moreover, the heights of players vary from several pixels to hundreds of pixels so that conventional deep neural networks often miss smaller players because the activation of smaller player vanishes at the end of a deep network. 

We propose a neural network to solve the problems raised above. We first design a cascaded convolutional neural network (CNN) for player/non-player classification. The cascaded CNN can quickly reject non-players in its shallow parts (i.e., early branches), greatly speeding up detection. We then present an end-to-end training approach to boost detection performance. Moreover, we apply a dilation strategy in testing to improve the detection accuracy. With these three techniques, our method can robustly detect players under dynamic camera views.

We are not the first to use cascaded CNNs~\cite{Anelia15}. However, our method is significantly different from previous methods. First, our method inherits feature maps from previous stages, which avoids recomputing the feature maps from the original image at every stage~\cite{Anelia15}. Second, our method jointly trains parameters in all stages via global optimization. It is significantly different from previous methods~\cite{Anelia15,FanYang16} in which cascade stages are relatively independent. In addition, the main purpose of the dilated strategy in our method is to align feature maps instead of increasing the receptive field~\cite{Dilation16}. 

The main contributions of this work are three-fold:

\begin{itemize}
	\vspace{-0.7ex}
	\item[\(\bullet\)] Design a cascaded CNN and a joint learning objective so that the learned model can quickly reject non-players in sports player detection. The trained model is very compact (less than 100KB) and is very efficient in testing (about 10 fps for images of 1280\(\times\)720 with un-optimized Matlab code);\vspace{-0.6ex}
	
	\item[\(\bullet\)] Present a dilation strategy to achieve accurate player detection on various conditions such as varying appearance, dynamic camera movements and complex background;\vspace{-0.6ex}    
	
	\item[\(\bullet\)] Propose a soccer player detection dataset {which is available on-line\footnote{\url{http://www.cs.ubc.ca/~jhchen14/ccnn_player_detection/}} for research purposes.} It contains numerous challenging situations such as varying illumination, player appearances, poses, zoom levels, motion blur, severe occlusions and cluttered backgrounds. This dataset is complementary to the APIDIS and SPIROUDOME datasets~\cite{APIDIS_ds,SPIROUDOME_ds} which focus on basketball games.\vspace{-0.7ex}
    
\end{itemize}

We evaluate our approach on basketball and soccer games. In all experiments, our method can accurately detect players under challenging conditions.

%\vspace{-2mm}
\section{Related work}
\label{sec:rel_works}

\paragraph{Player detection}
Player detection from sport video has been addressed by a wide variety of methods in recent years. The most common approaches are based on background subtraction~\cite{Chang90,Zhong04,carr2012monocular}. For instance, Zhong \etal~\cite{Zhong04} have introduced a domain-independent global color filtering method to extract player regions from background. In the same vein, Chang \etal~\cite{Chang90} first estimate the dominant color of the court and then detect player candidates from the background. These approaches are efficient, but their performance can be easily affected by illumination changes, camera movements and the presence of spectators~\cite{Manafifard17}. To make player detection more robust, researchers have developed stronger features such as histogram of oriented gradients (HOG) features ~\cite{HOG05} with the support vector machine (SVM) method~\cite{Mackowiak13,Baysal16}.  Moreover, researchers have combined different features such as edge~\cite{Weicun14}, LBP~\cite{LBP14} and motion~\cite{Schlipsing14} or have employed part-based model~\cite{Wu13,Ivankovic14} to improve performance. However, they are not as robust as deep learning based methods in general.\vspace{-2ex}

\paragraph{CNN-based object detection}
In recent years, deep learning has boosted the development of object detection. Convolutional neural networks (CNNs)~\cite{CNN12,walach2016learning} stands out as one of the most competitive methods for object detection. For example, R-CNN~\cite{RCNN12} first employs Selective Search~\cite{SelectiveSearch13} to generate candidate bounding boxes (object proposals) and then applies CNNs to classify objects from these proposals. Thereafter, Fast R-CNN~\cite{Fast_RCNN15} and Faster R-CNN~\cite{FasterRCNN16} have improved the performance of the region proposal based methods. Another set of approaches regard object detection as a regression problem. You only look once (YOLO)~\cite{YOLO15} and Single Shot MultiBox Detector (SSD)~\cite{SSD15} are two well-known regression-based methods. Both of them can  simultaneously output the bounding box, category and confidence score for each detected object. However, they require a large-scale CNN and yet their performance on small objects detection is unsatisfactory. To improve the performance of CNN-based methods, cascaded CNNs have been proposed to eliminate non-object samples step by step. Li \etal~\cite{Haoxiang15} have introduced a multi-resolution CNN cascade to quickly reject background regions for face detection. Angelova \etal~\cite{Anelia15} have used cascade deep nets and fast features for efficient pedestrian detection. More recently, Yang \etal~\cite{FanYang16} have constructed a cascaded architecture by applying discrete AdaBoost~\cite{Freund19} after each convolutional layer. 

 % introduction and related work
%\vspace{-2mm}
\section{End-to-end cascaded CNN}
\label{sec:CRC}

\begin{figure}[t]
	\centering
	\includegraphics[width=0.9\textwidth,keepaspectratio]{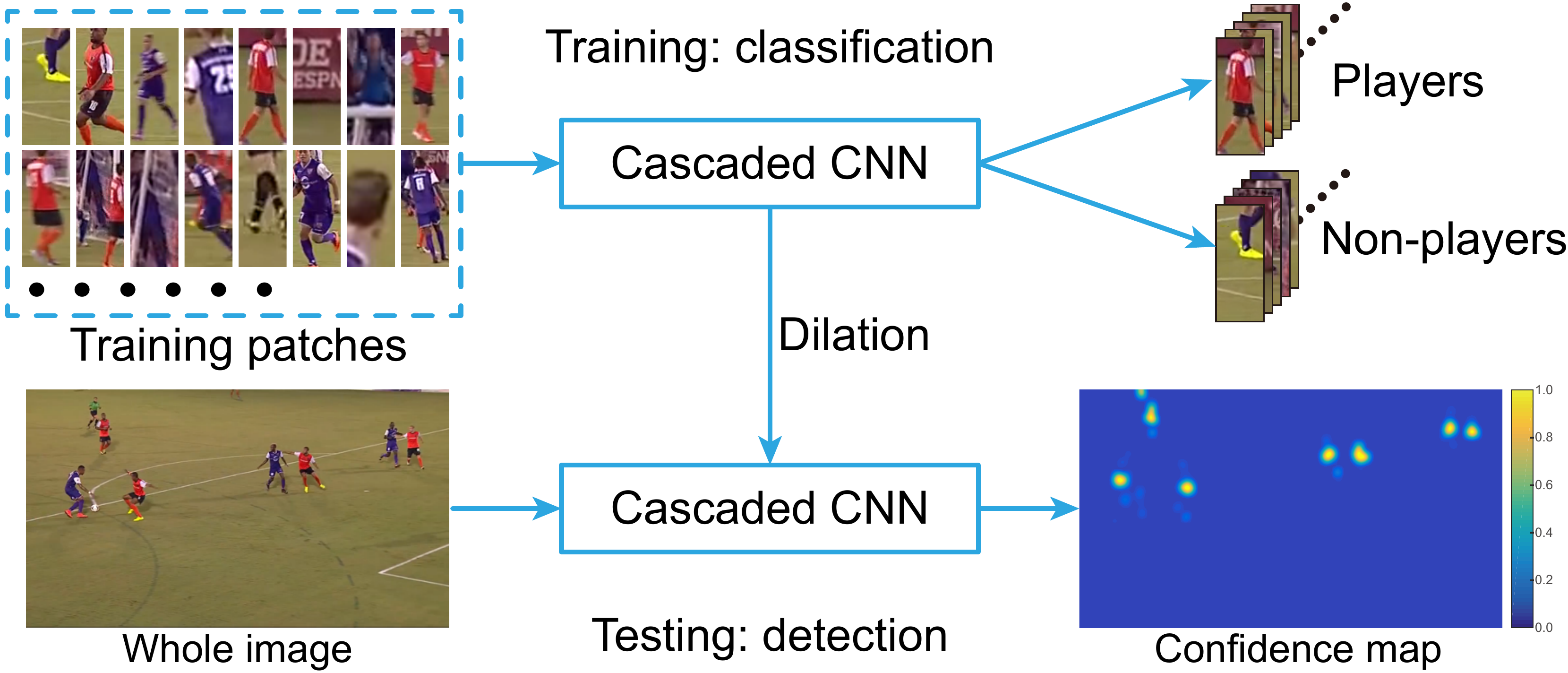}
	\vspace{-2.6ex}
	\caption{\textbf{Player detection pipeline}. We first train a neural network classifier from labeled image patches. Then, we detect player locations from a whole image using a dilation strategy. The network architecture, the training method and the dilation strategy are in Section \ref{sec:netarch}, \ref{sec:training} and \ref{sec:dilation}, respectively.}
	\label{fig:pipeline}
\end{figure}
\begin{figure}[t]
	\centering
	\includegraphics[width=0.9\textwidth,keepaspectratio]{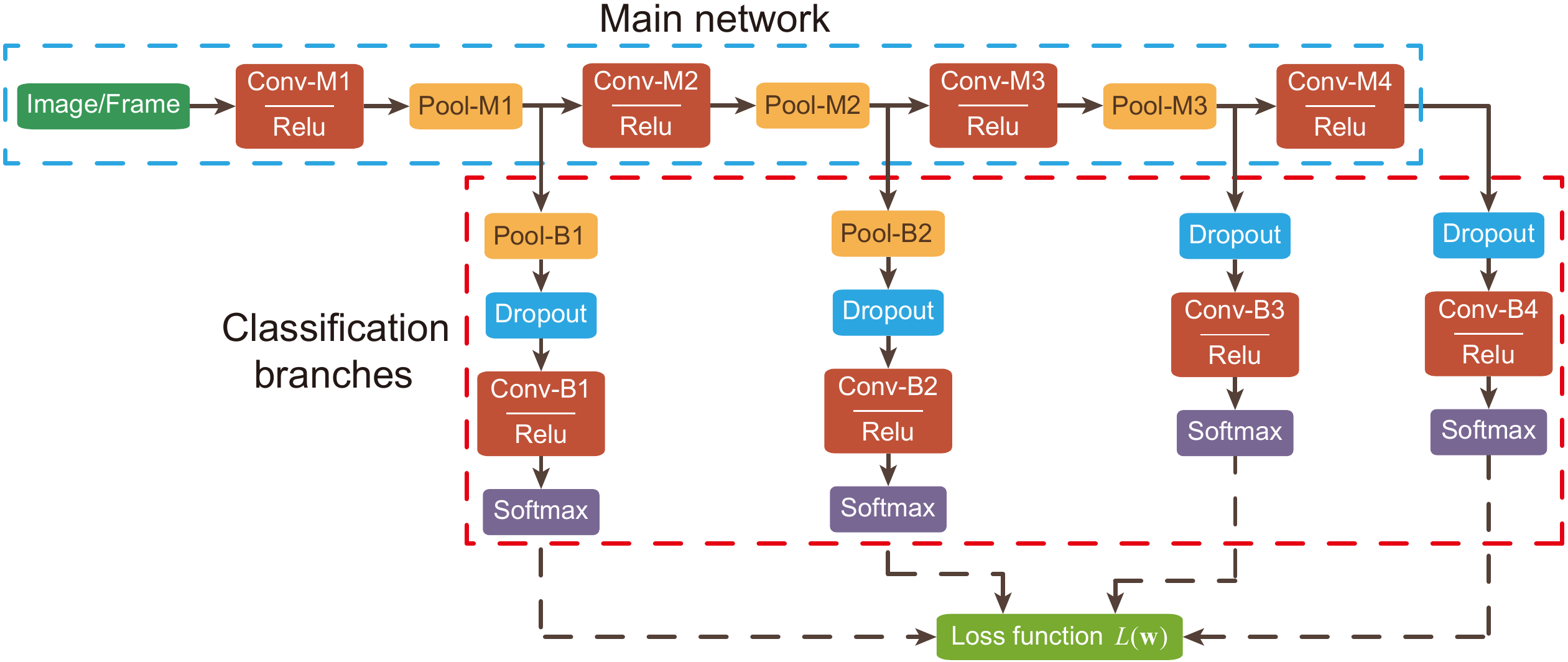}
	\vspace{-2.6ex}
	\caption{\textbf{Architecture of our cascaded CNN.} It has main network (blue box) and four branches (red box). The four branches are ordered (from left to right) to efficiently detect players. ``-Mx'' and ``-Bx'' stand for the x-th convolution or pooling layer in main network and classification branch, respectively.}
	\label{fig:architecture}
\end{figure}

Figure \ref{fig:pipeline} shows the pipeline of our method. It consists of a cascaded CNN for player/non-player classification, an end-to-end training approach for global optimization and a dilation strategy for accurate detection.

\subsection{Neural network architecture}
\label{sec:netarch}

Our neural network has a cascaded architecture (See Figure \ref{fig:architecture}). It contains a main network (dashed blue box) and four classification branches (dashed red box). The main network is modelled after AlexNet \cite{CNN12}. However, we use much fewer filters (16 or 32 \textit{vs} up to $2,048$) with a small size ($3 \times 3$) in each layer. Each branch is a shallow and simple network that classifies if the input has a player or not. As a result, the whole network is very light (less than 100 KB) compared with a conventional network which usually is more than 100 MB.

Our cascaded architecture fits the player detection task very well. For example, let us assume the input is an image patch. It will be passed to classification branches in order. If and only if the output of previous branch is positive (i.e., above an threshold), the image patch will go to next branch. By doing so, our network has two advantages. First, most negative examples are eliminated in the earlier branches so that our network is computationally efficient. Second, each branch can be trained for different levels of hardness of player detection. For example, branch one can eliminate most easy negative examples such as pure playing ground and leave hard examples to later branches. On the other hand, later branches (like branch four) can be specifically trained using hard examples. As a result, the whole network is more powerful when distinct branches work together. 

Our network has an additional benefit that feature maps are partly shared in the main network. Because the main network and the classification branches are connected in a unified CNN framework,  we can perform operations on the feature maps constructed by previous stage instead of sampling image patches from the original image. 

\subsection{Network training}
\label{sec:training}

The training procedure has two steps: branch-level training and whole network training. The first step is crucial as it can significantly speed up the convergence rate of subsequent training process. In branch-level training, we regard each branch as an independent network. When the training of a particular branch is completed, we choose a recall rate threshold (97\%) to remove negative samples that will not be used to train later branches. The four branches are trained in order using different sets of labeled samples.

When the branch-level training is completed, we perform end-to-end network training that simultaneously optimizes all the cascade branches. Our cascaded CNN is a cascaded classifier. Assume that the model has \(K\) cascade stages and is trained on \(N\) image samples. Let \(D = \{ ({x_{i,j}},{y_{i,j}})\} \) denote the training set where \(1 \le i \le N\) and \(1 \le j \le K\). \({x_{i,j}} \in {\mathbb{R}^d}\) is the feature map of the \(i\)-th sample at the \(j\)-th cascade stage and \({y_{i,j}} \in \{ 0,1\} \) is the corresponding binary label. Then the probability for a sample to be positive is:
\begin{equation}
{
	{p_{i}}({{y}_i} = 1|{{\mathbf{x}}_i},{\mathbf{w}}) = \prod\limits_{j = 1}^K {{p_{i,j}}({y_{i,j}} = 1|{x_{i,j}},{\bf{w}})},
}
\label{eq_likehood_p}
\end{equation}
where \({\bf{w}}\) is the weights of the cascaded CNN. If a sample is classified as negative by any one of the cascade stages, our method predicts it as negative. Accordingly, we have:
\begin{equation}
{
	{p_i}({y_i} = 0|{x_i},{\bf{w}}) = 1 - \prod\limits_{j = 1}^K {{p_{i,j}}({y_{i,j}} = 1|{x_{i,j}},{\bf{w}})}.
}
\end{equation}

Inspired by~\cite{CascadeOpt10}, the loss function can be defined as:
\begin{equation}
{
	{L_p}({\bf{w}}) =  - \sum\limits_{i = 1}^N {\left[{{y_i}\log ({p_i}({y_i} = 1|{x_i},{\bf{w}})} ) + (1 - {y_i})\log ({p_i}({y_i} = 0|{x_i},{\bf{w}}))\right]}.
}
\label{eq_loss_p}
\end{equation}

In equations (\ref{eq_likehood_p})$\sim$(\ref{eq_loss_p}), which stage first predicts an example as negative (negative elimination) does not affect the final result. However, fast negative elimination affects computational cost. For instance, if more negative samples are rejected by earlier cascade stages, fewer samples will be passed to upper stages, reducing the overall computational cost. 

To achieve fast negative elimination, we design a regularization term for the loss function. Let ${T_j}$ be the computational cost of the $j^{th}$ cascade stage, which can be estimated according to the sizes of input feature maps, pooling and convolution kernels in this stage. Then, the regularization term is:
\begin{equation}
{
{L_\Gamma }({\bf{w}}) = \frac{1}{N}\sum\limits_{i = 1}^N {\sum\limits_{j = 1}^K {{T_j}}  \cdot \left( {\prod\limits_{u = 1}^j {{p_{i,u}}({y_{i,u}} = 1|{x_{i,u}},{\bf{w}})} } \right)}.
}
\end{equation}

The final loss function is a weighted function of accuracy cost and computation cost:
\begin{equation}
{
	L({\bf{w}}) = {L_p}({\bf{w}}) + \beta {L_\Gamma }({\bf{w}}),
}
\label{equ:joint_loss}
\end{equation}
where $\beta$ is a weight to balance the accuracy term and regularization term. In this work, $\beta$ is experimentally set to 0.5.

With this loss function, we train the whole cascaded CNN end-to-end using Adam algorithm~\cite{Adam15}. Then, we estimate the cascade thresholds \({\bf{\lambda }} = \{ {\lambda _1}, \ldots ,{\lambda _K}\} \) using a grid search over a range of thresholds. Only samples that satisfy \({p_{i,j}}({y_{i,j}} = 1|{x_{i,j}},{\bf{w}}) > {\lambda _j}\) can pass the $j^{th}$ cascade stage.

\subsection{Dilation strategy for accurate detection}
\label{sec:dilation}
\begin{figure}[t]
	\centering
	\includegraphics[width=0.9\textwidth,keepaspectratio]{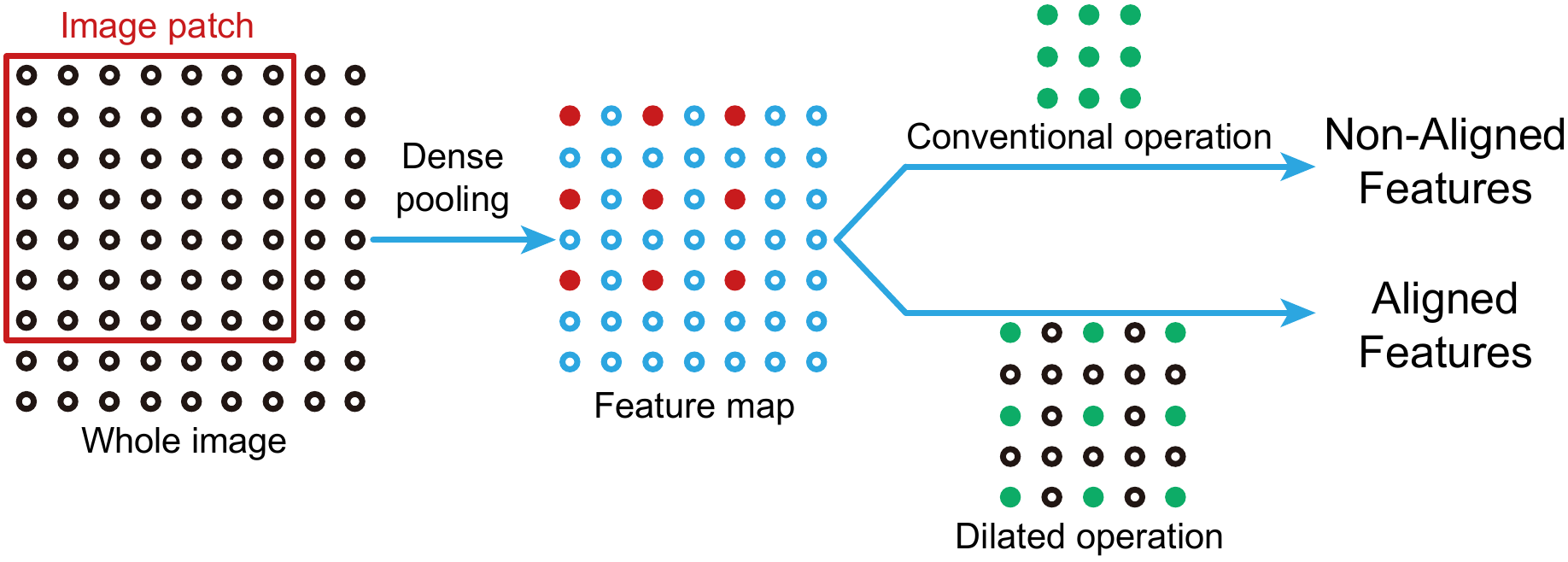}
	\vspace{-2.6ex}
	\caption{\textbf{The motivation of using the dilation strategy.} In the feature map (middle one) produced by dense pooling (trained with \({\rm{stride = 2}}\)), the feature points belonging to the red image patch are marked with red, which are separated by feature points generated by its neighbor image patches (marked with blue). In dilated operation (bottom right), valid positions (green points) are separated so that they properly align with the feature map generated by the previous layer.}
	\label{fig:dilation}
\end{figure}

Because the neural network introduced above is trained using image patches, directly applying it to a whole image would generate unaligned feature maps for operation kernels (see Figure \ref{fig:dilation}). To address this issue, we develop a dilation strategy.

The dilation strategy aims to align feature maps before and after convolution and pooling layers in testing. Let us use the pooling layer as an example. In training, the pooling layer is  $3 \times 3$ with ${\rm{stride}} = 2$. In testing, the stride of the pooling layer should be changed to 1 to obtain an accurate feature location. As a result, the feature map will be misaligned before and after pooling (see Figure \ref{fig:dilation}, top right), resulting in incorrect feature maps or requiring post-processing.

To address this problem, we employ a dilation strategy like \cite{Dilation16} for convolution and pooling. For simplicity, we use 2D convolution as an example. Let $w$ denote the convolution kernel (or weight) with size $m \times n$, each convolution step can be written as:
\begin{equation}
{
\eta  = \sum\limits_{i = 1}^m {\sum\limits_{j = 1}^n {{x_{i,j}} \cdot {w_{i,j}}} } ,
}
\end{equation}
where \(x\) is the corresponding 2-D data of this convolution step. Let \({\hat w}\) and \(l\) be the dilated convolution kernel and dilation factor, respectively. The dilated version of convolution can be formulated as:
\begin{equation}
{
\hat \eta = \sum\limits_{i = 1}^m {\sum\limits_{j = 1}^n {{x_{i,j}} \cdot {{\hat w}_{(i - 1) \cdot l + 1,(j - 1) \cdot l + 1}}} } .
}
\end{equation}

In the same way, dilated pooling can be performed by using the kernel with valid points located at \(((i - 1) \cdot l + 1,(j - 1) \cdot l + 1)\). In dilated operation, as shown in Figure \ref{fig:dilation}, valid positions are separated and the intervals of these positions are controlled by the dilation factor \(l\). Consequently, the dilated kernel can spatially align with the separated feature map from dense pooling by adjusting the dilation factor. In this way, our neural network can be applied to the whole image without sacrificing detection accuracy. Moreover, the dilation strategy improves the efficiency of the method because it makes feature map sharing available in the whole image. More details are provided in the supplementary material.

%\vspace{-2mm}
\section{Experiments}
\label{sec:experiment}

\subsection{Datasets and error metrics}
\label{subsec_dataset}

\begin{figure}[t]
	\centering
	\subfigure[Players' heights]{\includegraphics[height=18ex,keepaspectratio]{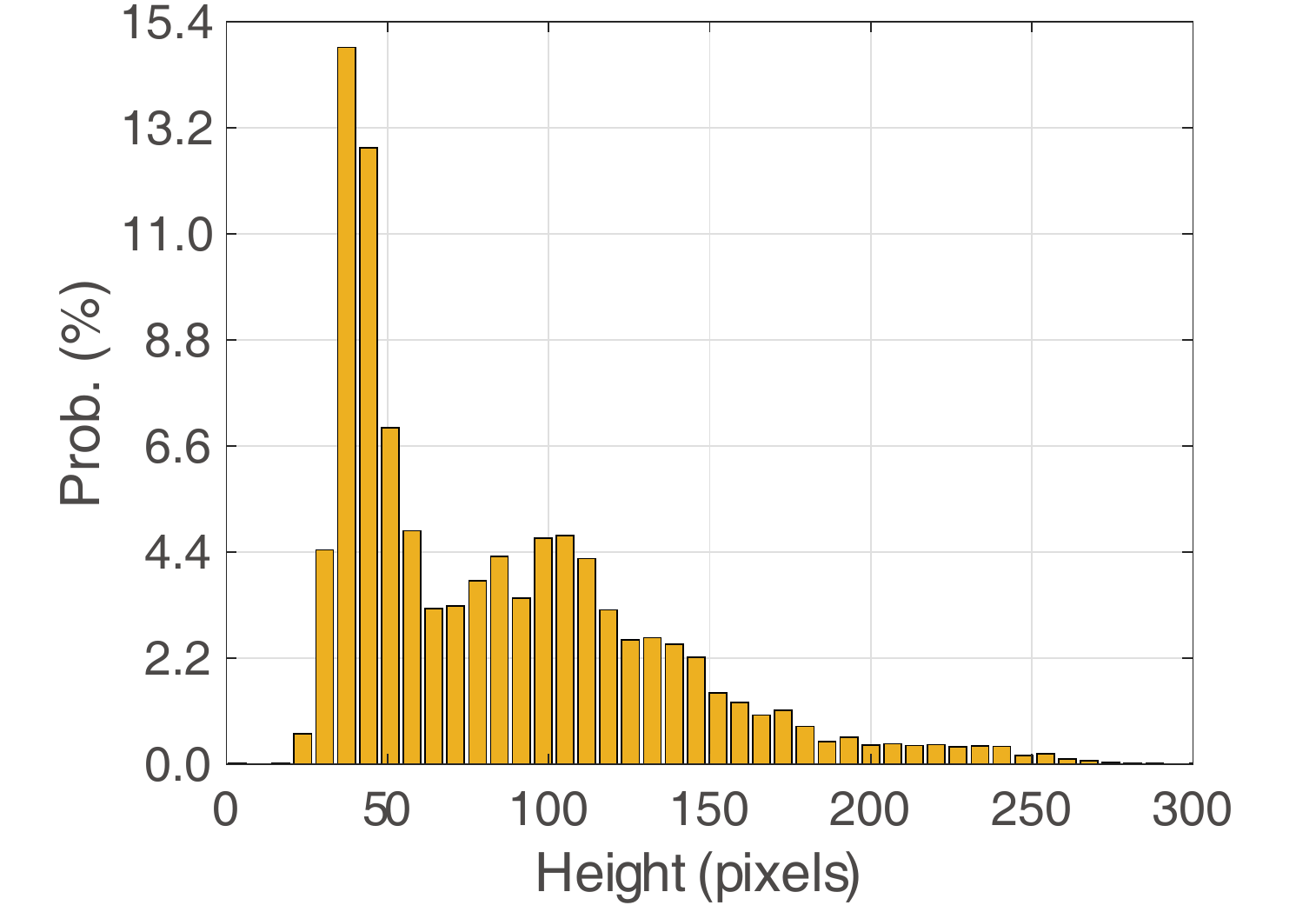}
		\label{fig_PlayerHeight}}
	\subfigure[Players' centers]{\includegraphics[height=18ex,keepaspectratio]{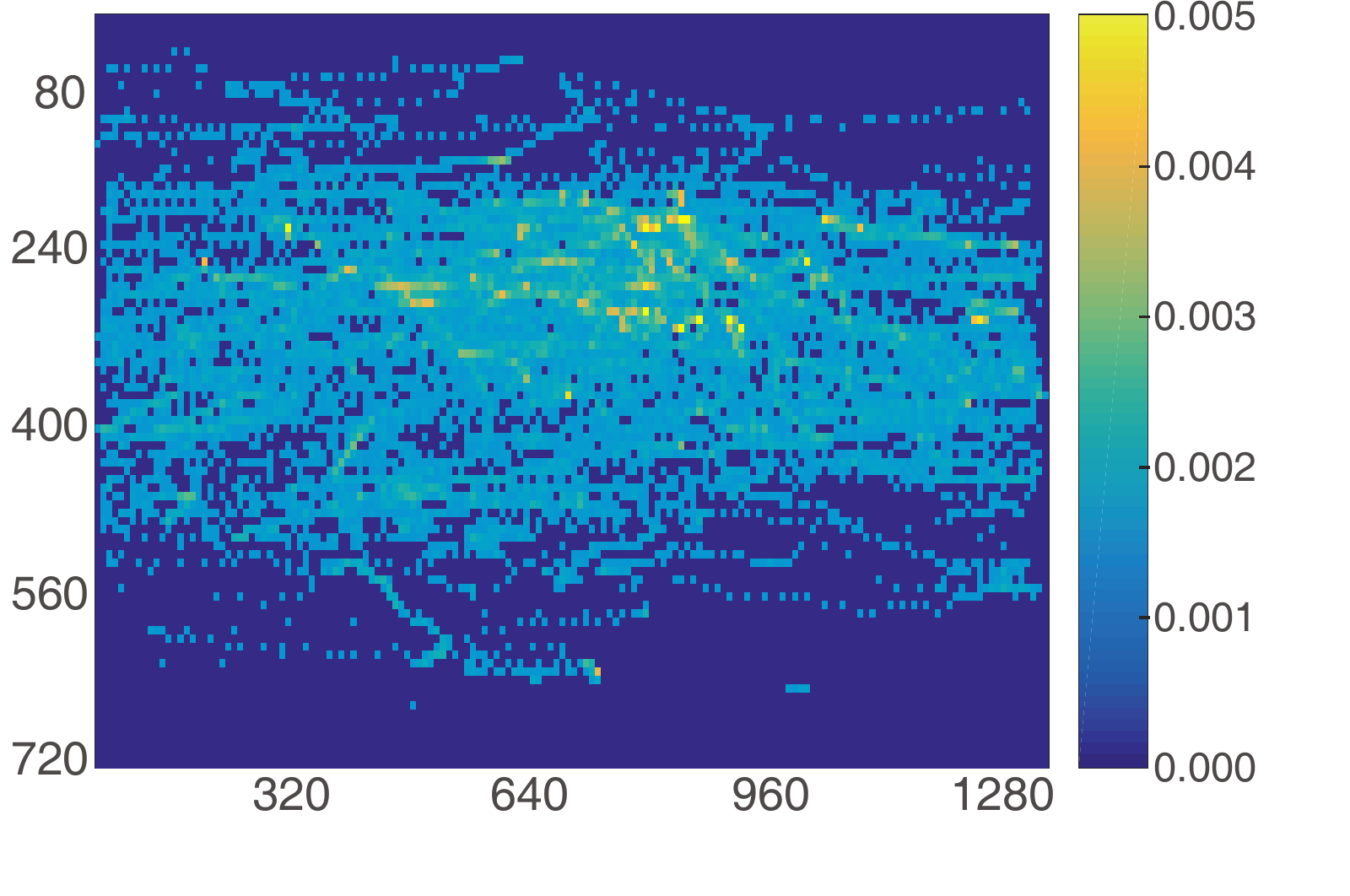}
		\label{fig_Distri_Center}}
	\subfigure[Number of players per frame]{\includegraphics[height=18ex,keepaspectratio]{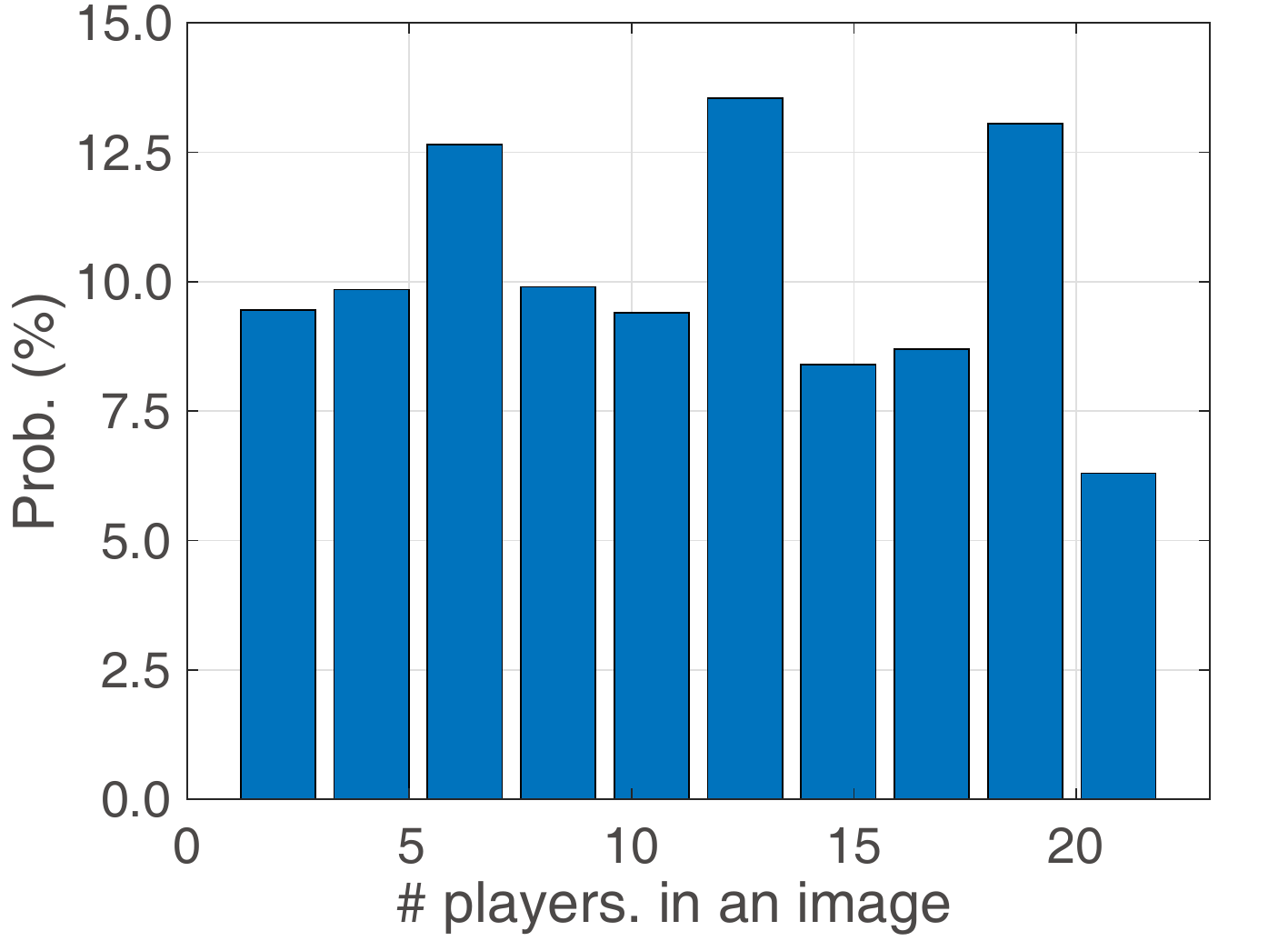}
		\label{fig_NumPerImage}}
	%\caption{Statistics on our soccer dataset.}
	\vspace{-2.6ex}
	\caption{\textbf{Statistics of our soccer dataset.} The figures show the distributions of players' heights, centers and number of players per frame. }
	\label{fig_statistics}
\end{figure}

\paragraph{Soccer dataset}
The soccer dataset is created from two professional matches hosted in the same stadium. Each match was recorded by three broadcast pan-tilt-zoom (PTZ) cameras (30 FPS and $1280 \times 720 $ resolution). One camera is located at mid-field, overlooking the game. The other two cameras are located behind the left and right goal gates separately. Highlight video sequences were selected from the original video by professional editors. The highlight video represents typical soccer games events such as goals, goal attempts, passings and penalty kicks. 22,586 player locations were manually annotated from 2,019 images. The dataset contains numerous challenges such as varying player appearances, poses, zoom levels, motion blur, severe occlusions and cluttered background. Figure \ref{fig_statistics} shows the statistics of the dataset. The height of players, the image location of players and the number of players per image are widely distributed, demonstrating the diversity of the dataset. For example, the height of players is from about 20 pixels to 250 pixels with a long tail distribution from height of 150 pixels.\vspace{-2ex}

\paragraph{Basketball dataset}
We use two standard basketball datasets: APIDIS ~\cite{APIDIS_ds} and SPIR-OUDOME ~\cite{SPIROUDOME_ds}. They were originally used for intelligent basketball video analysis ~\cite{Chen10,Parisot2017}. In these datasets, the challenges are from lower color contrast between players and background, and highly dynamic player movements.

In both datasets, $50\%$ of images are randomly selected as training samples and the rest are used as testing data. For error metrics, we use an intersection-over-union (IoU) threshold of 0.7 to determine the correctness of detection.

\subsection{Comparison with baselines}
\label{subsec_compare_conv}
We build three baselines for the purpose of comparison. \textbf{Baseline 1} is a conventional CNN without a cascaded design. Table \ref{tbl_resource_com} top row (conventional) shows the structure of this baseline. \textbf{Baseline 2} is our method without end-to-end learning (i.e., with only branch-level training). \textbf{Baseline 3} is our method without the dilation strategy (Section \ref{sec:dilation}).

\begin{table}[t]
	%\footnotesize
	\scriptsize
	\centering	
    \scalebox{0.9}{
	\begin{tabular}{c | c c c c | c | c}
		\toprule 
		\multirow{2}{*}{Type} & \multicolumn{4}{c|}{Network Structure} & \multirow{2}{*}{Memory} & \multirow{2}{*}{AUC} \\	
		  & Conv-M1/B1 & Conv-M2/B2 & Conv-M3/B3 & Conv-M4/B4  &  &  \vspace{-0.4ex} \\
		\midrule
		 \multirow{4}{*}{Conventional} & 64 / -  & 128 / - & 256 / - & 512 / 2  & 5.95 MB & 0.873 \\
		 & 128 / -  & 256 / - & 512 / - & 1024 / 2 &  23.72 MB & 0.937 \\
		 & 256 / -  & 512 / - & 1024 / - & 1024 / 2  & 58.61 MB & 0.962 \\
		 & 256 / -  & 1024 / - & 1024 / - & 2048 / 2  & 81.10 MB & 0.977 \\
		\midrule
		 \multirow{4}{*}{Cascaded (Ours)} & 8 /2  & 8 / 2 & 8 / 2 & 8 / 2  & \textbf{12.46 KB} & 0.881 \\
		 & 8 / 2  & 16 / 2 & 16 / 2 & 16 / 2  & \textbf{31.84 KB} & 0.932 \\
		 & 16 / 2  & 16 / 2 & 16 / 2 & 16 / 2 & \textbf{38.18 KB} & 0.967 \\
		 & 16 / 2  & 16 / 2 & 32 / 2 & 32 / 2 & \textbf{79.43 KB} & 0.973 \\
		\bottomrule
	\end{tabular}
    }
    \vspace{-2.6ex}
    \caption{\textbf{Comparison between a conventional CNN and our cascaded CNN.} Our design achieves about $1000\times$ memory (storage of network weights by single-precision floating-point) saving without sacrificing prediction performance. In this table, \doubleQuote{Conv-Mx/Bx} stands for the number of filters in the x-th convolutional layer in the main network and its classification branch, respectively. All these filters have the size of \(3 \times 3\). Detection performance is measured by area under curve (AUC) of receiver operating characteristic (ROC) curve on the soccer dataset.}
    \label{tbl_resource_com}
\end{table}
%Memory is measured by the weights of networks (single-precision floating-point).

Table \ref{tbl_resource_com} shows the comparison of our method with baseline 1. Our method is able to achieve a similar performance by using much less memory consumption (about $1000 \times$). Although baseline 1 has no classification branches, it requires a large network to detect players from challenging scenarios. On the contrary, our method uses a cascaded network to reduce the complexity of the player detection problem and thus outperforms baseline 1 in terms of memory consumption. Table \ref{table:cascade} shows the performance of our method in all stages. The accuracy is improved in every stage, especially from stage 1 to stage 2.

\begin{figure}[t]
	\begin{floatrow}
		\ffigbox{%
			\includegraphics[width=0.39\textwidth,keepaspectratio]{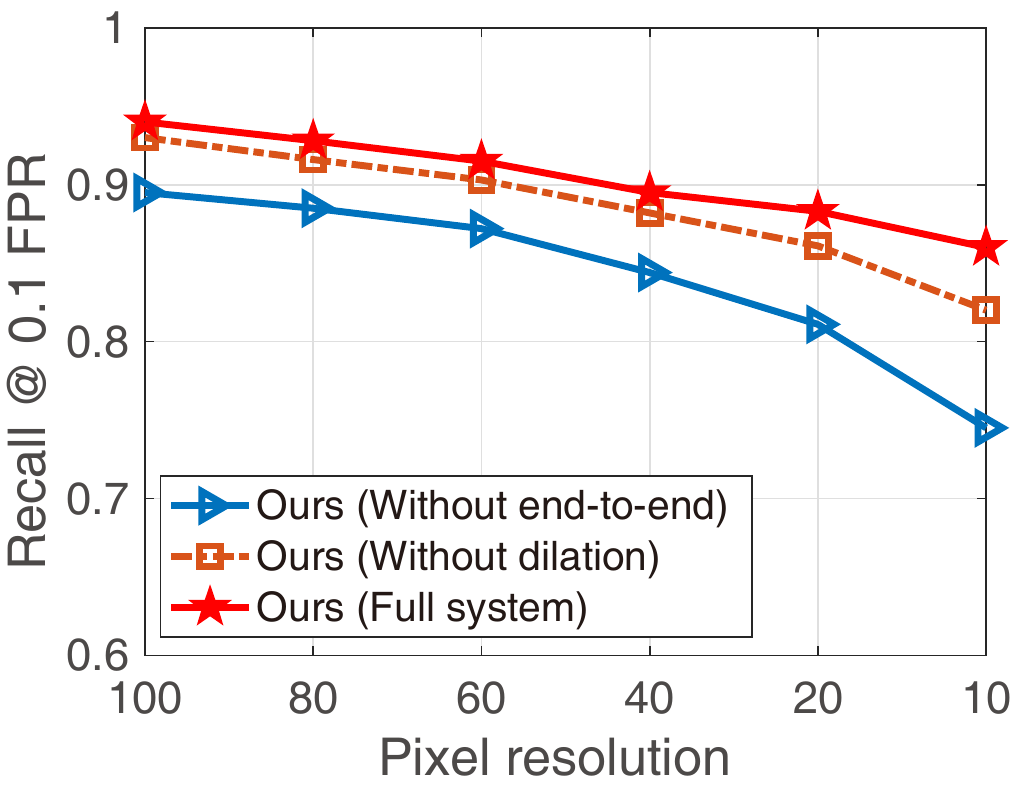}
		}
		{%
		\vspace{-2.6ex}
		\caption{\textbf{Influence of the end-to-end learning and the dilation strategy.} It shows the recall @ 0.1 false positive rate (FPR) as a function of the players' pixel resolution on the soccer player dataset.}
		\label{fig:wihout_end_to_end}
		}
		\ffigbox{%
			\includegraphics[width=0.39\textwidth,keepaspectratio]{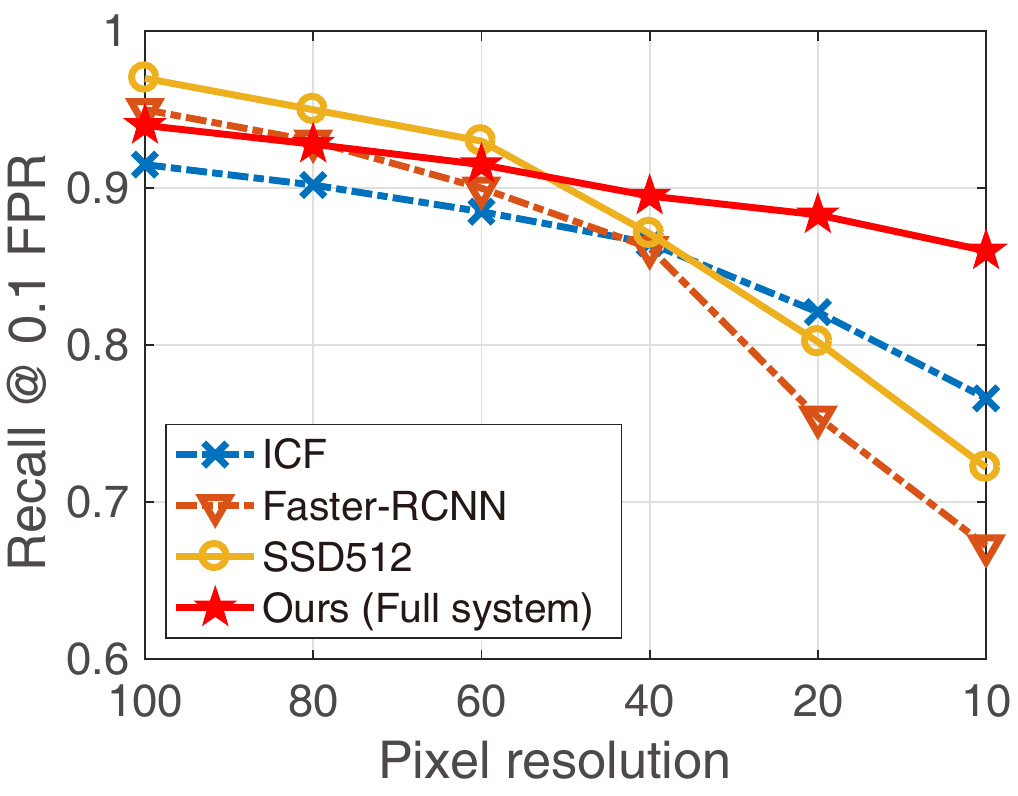}
		}{%
		\vspace{-2.6ex}
	\caption{\textbf{Recall with different pixel resolutions.} The plot shows recall rate @ 0.1 false positive rate (FPR) as a function of players' pixel resolution on the soccer dataset.}
		\label{fig_Reso}
	}	
\end{floatrow}
\end{figure}

Figure \ref{fig:wihout_end_to_end} shows the comparison of our method with the baseline 2 and the baseline 3 on the soccer dataset. Our method substantially improves the recall rate (at 0.1 false positive rate) in all levels of player resolutions. 
%\vspace{-2mm}
\subsection{Comparison with state-of-the-art methods}

\begin{figure}[t]
	\begin{floatrow}
		\capbtabbox{%
			\scriptsize
			\begin{tabular}{p{12pt}<{\centering}|c|p{23pt}<{\centering}}
				\toprule
				\multirow{2}{*}{Stage} &  Recall & \multirow{2}{*}{Accuracy}  \\ 
				 &  @ 0.1 FPR & \\
				\midrule   
				1 & 0.991 & 0.463 \\
				2 & 0.973 & 0.872 \\
				3 & 0.967 & 0.931 \\
				4 & 0.962 & 0.944\\
				\bottomrule
			\end{tabular}
		}
		{%
			\vspace{-2.6ex}
			\caption{\textbf{Performance of all the stages on soccer dataset.} Non-player samples are rejected stage by stage.}
			\label{table:cascade}  
		}
		\capbtabbox{%
			\scriptsize
            \scalebox{0.9}{
			\begin{tabular}{p{42pt} | p{12pt}<{\centering} p{20pt}<{\centering} p{12pt}<{\centering} | p{12pt}<{\centering} p{12pt}<{\centering} p{20pt}<{\centering} p{12pt}<{\centering}}
				\toprule				
				& \multicolumn{3}{c|}{Cross game}  &    \multicolumn{4}{c}{Same game}            \\
				
				Dataset     &ICF& Faster RCNN &  Ours & BRF   & ICF   &  Faster RCNN & Ours            \\
				
				\midrule
				APIDIS        &0.838& 0.887&  \textbf{0.910} & 0.950 & 0.968 &  0.941       & \textbf{0.976}  \\
				SPIROUDOME    &0.789& 0.854&  \textbf{0.889} & 0.949 & 0.951 &  0.923       & \textbf{0.965}  \\
				Soccer Set 1  &0.802& 0.850&  \textbf{0.901} & -     & 0.918 &  0.938       & \textbf{0.971}  \\
				Soccer Set 2  &0.815 & 0.867&  \textbf{0.908} & -     & 0.925 &  0.931       & \textbf{0.969}  \\
				\bottomrule
			\end{tabular}
            }
		}
		{%
			\vspace{-2.6ex}
			\caption{ \textbf{Performance comparison with state-of-the-art methods.} The detection performance is measured by areas under curve (AUC) of ROC curves. The best performance is highlighted.}
			\label{tbl_compare1}
		}
	\end{floatrow}
\end{figure}

We compare the proposed approach with several state-of-the-art algorithms: BRF~\cite{Parisot2017}, ICF~\cite{Int09}, Faster-RCNN~\cite{FasterRCNN16} and SSD512~\cite{SSD15}. BRF~\cite{Parisot2017} is a scene-specific classifier that is specifically designed for sport players detection. ICF~\cite{Int09} is a popular approach using hand-crafted features and the technique of integral channel features. Faster-RCNN~\cite{FasterRCNN16} and SSD512~\cite{SSD15} both adopt CNNs for object detection, but the difference is that the former is a region proposal based method while the latter is a regression based method.

We conduct experiments on both the soccer dataset and the basketball dataset. To evaluate generalization capacity of these methods in player detection, we did cross game evaluation in which the model is trained on one game and tested on a different game without fine-tuning. There results are denoted by an extra ``$-CRS$'' (e.g.\ ICF$-CRS$).

\begin{figure}[t]
	\centering
	\subfigure[Test on APIDIS]{\includegraphics[width=0.36\textwidth,keepaspectratio]{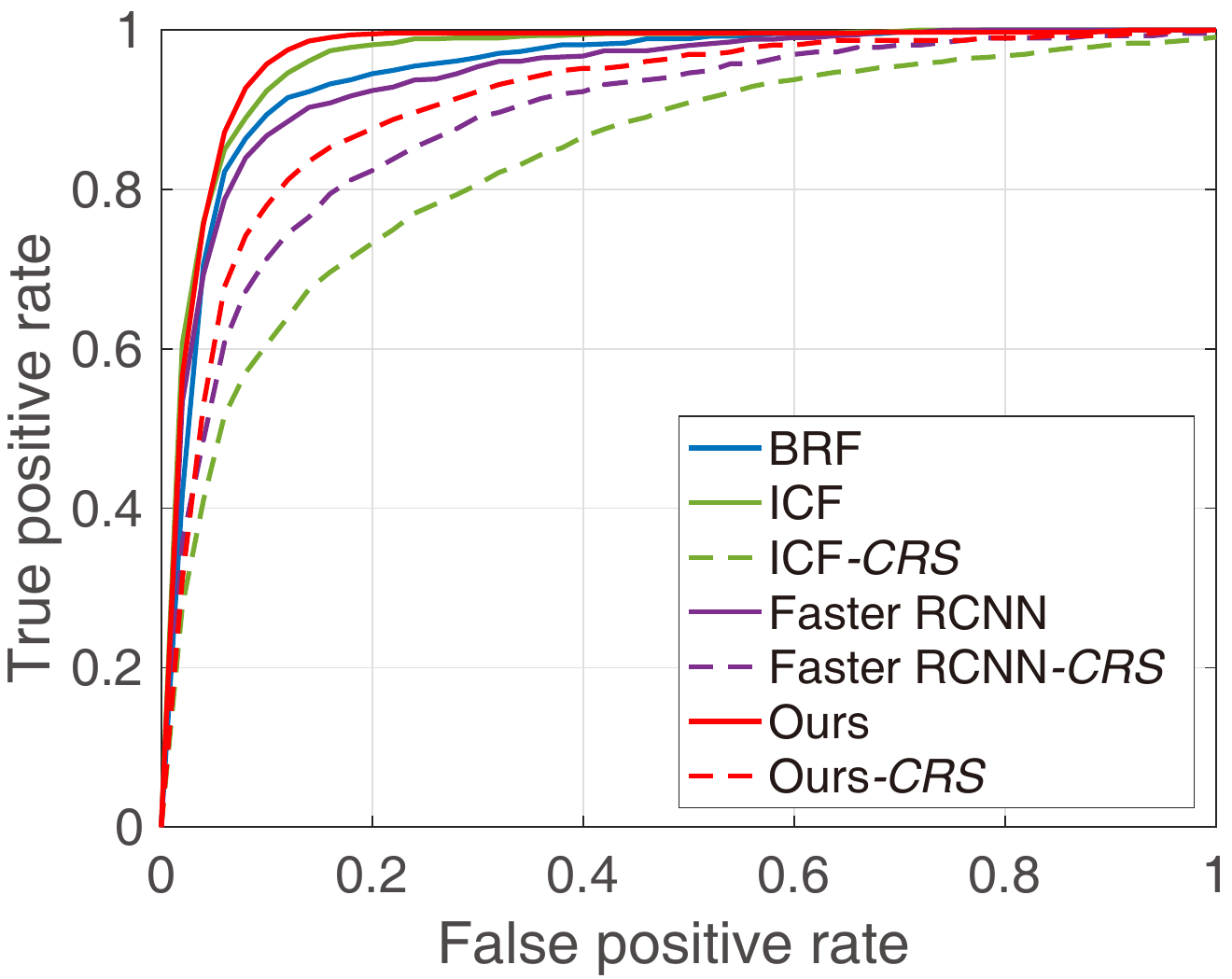}}
	\subfigure[Test on SPIROUDOME]{\includegraphics[width=0.36\textwidth,keepaspectratio]{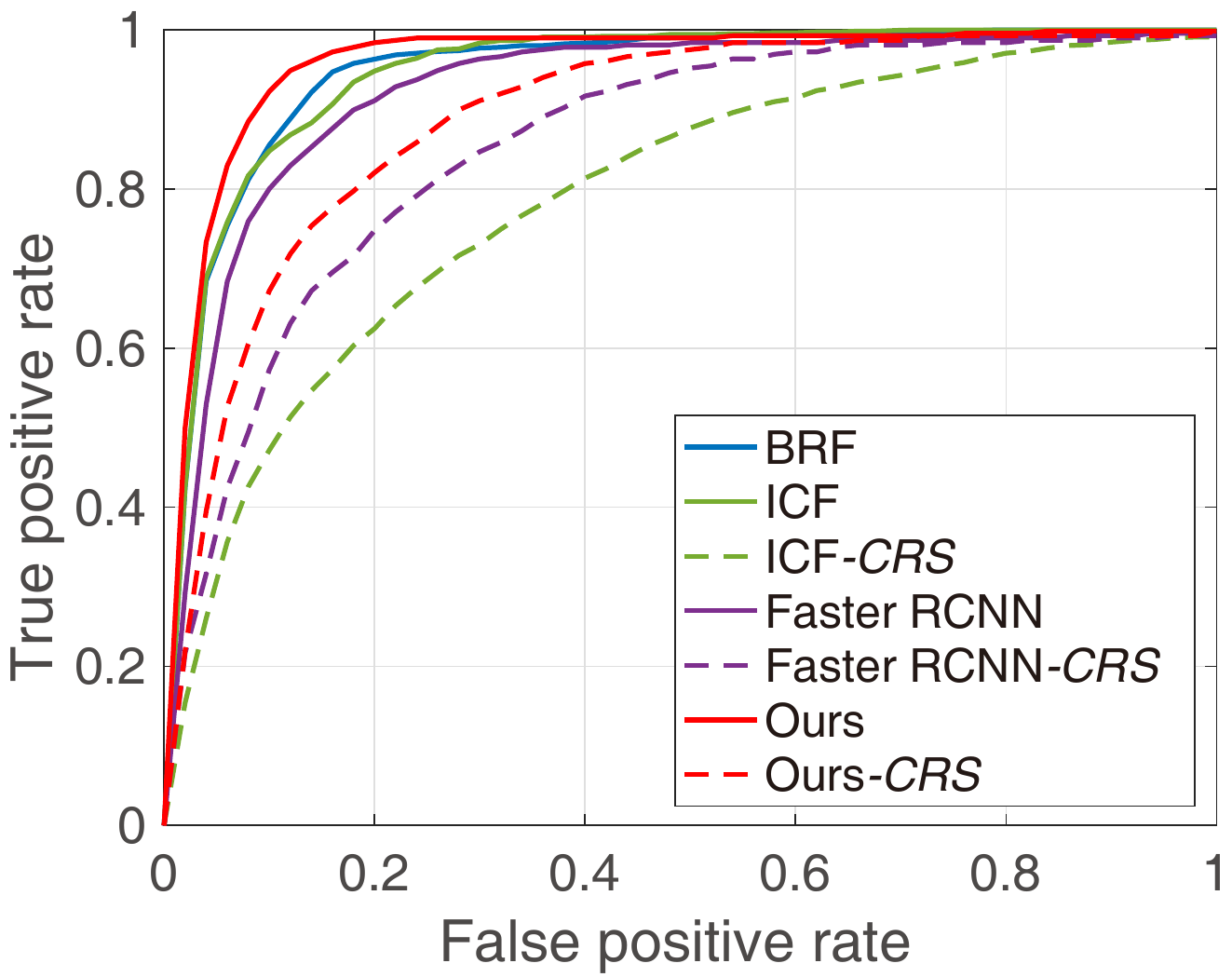}}
	\subfigure[Soccer Set 1]{\includegraphics[width=0.36\textwidth,keepaspectratio]{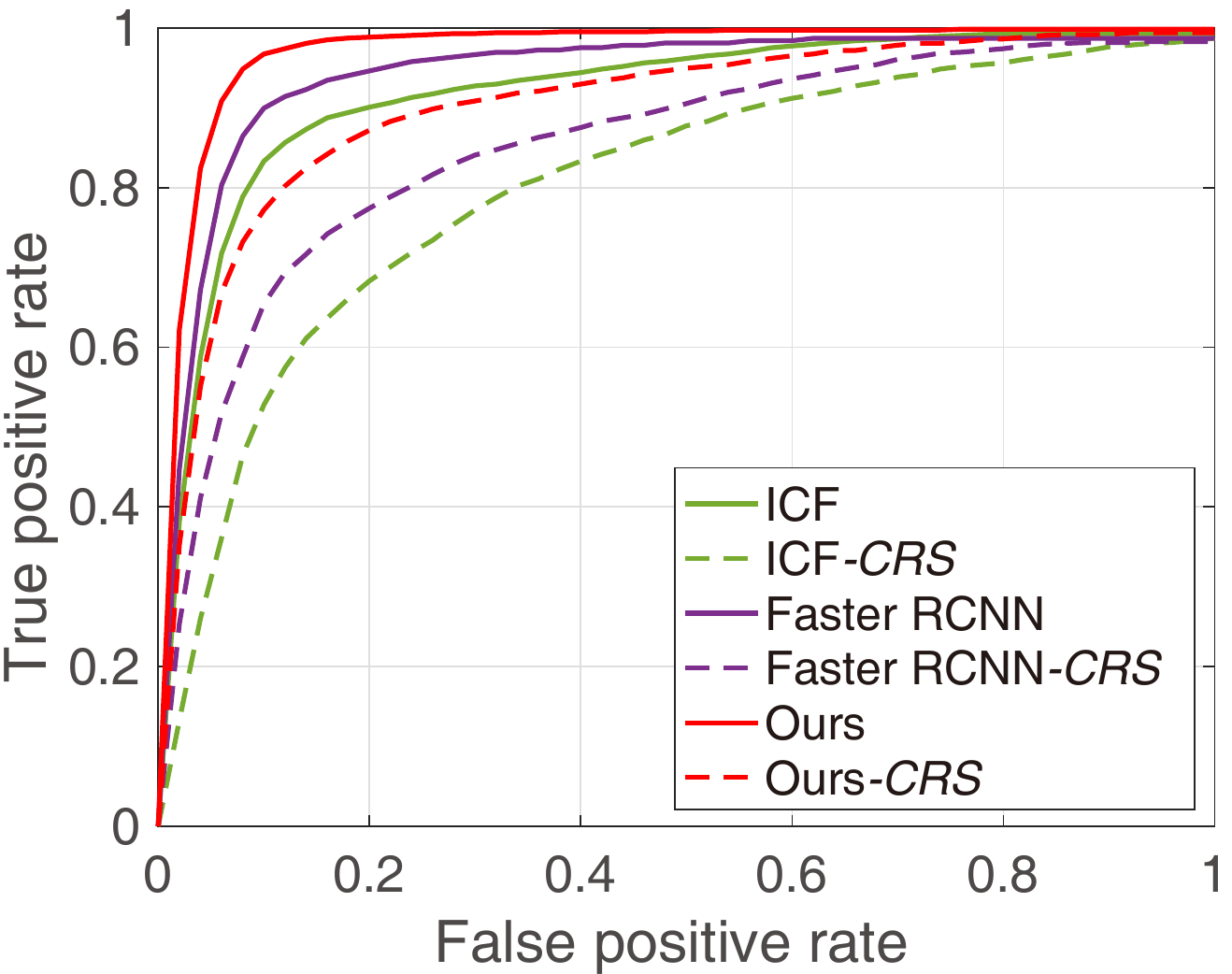}}
	\subfigure[Soccer Set 2]{\includegraphics[width=0.36\textwidth,keepaspectratio]{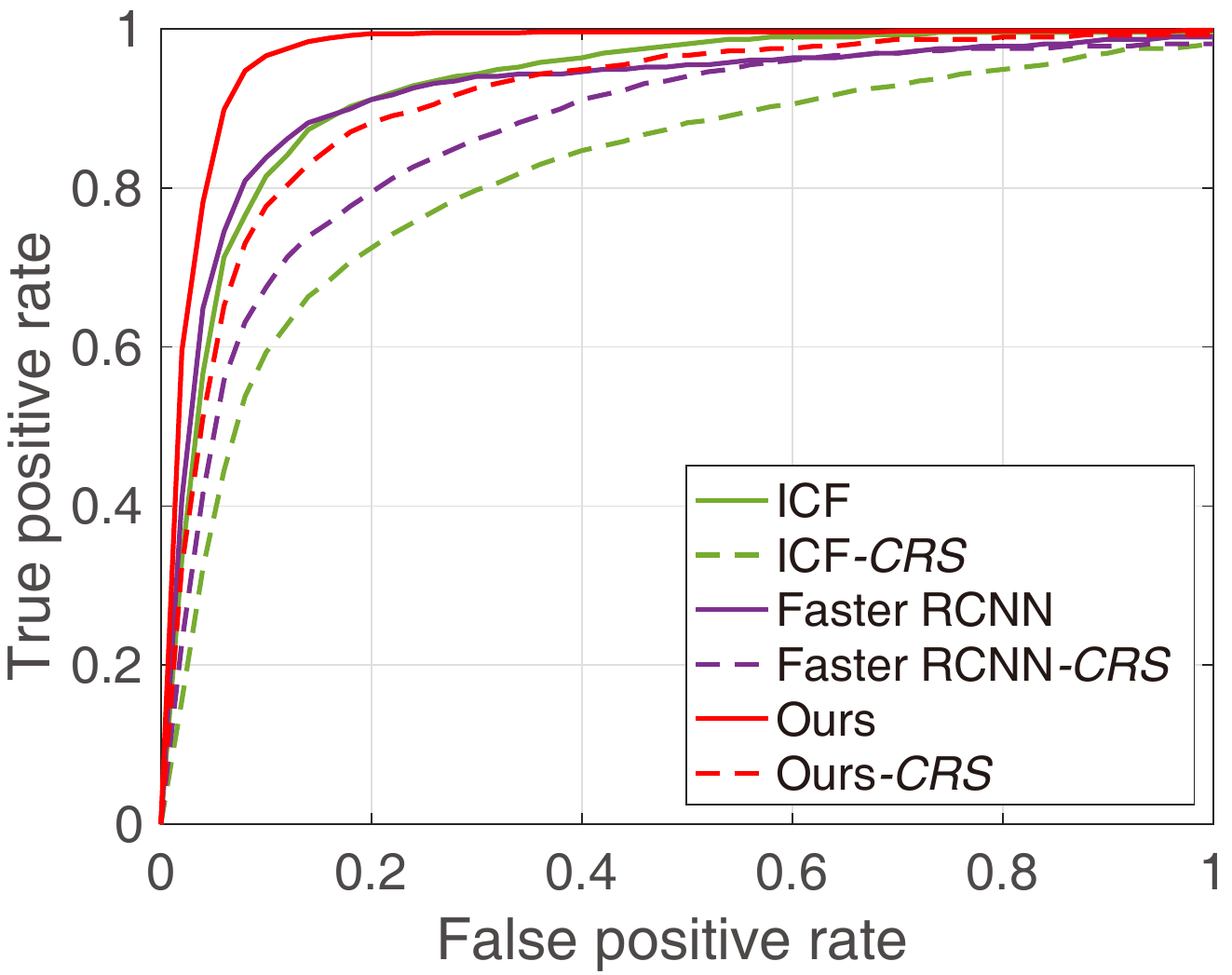}}    
	\vspace{-2.6ex}
	\caption{\textbf{Performance evaluation.} These figures show the ROC curves of each method on player detection benchmarks. The curve closer to top left represents better performance. }    
	\label{fig_rst_comp}
\end{figure}

Figure \ref{fig_rst_comp} shows the receiver operating characteristic (ROC) curves of all the methods. For both datasets, our method achieves the best performance, with the performance gap being especially pronounced in soccer. In the cross games evaluation, our method (red-dash lines) is also better than other methods, indicating higher generalization capability of our approach. The plotted lines of SSD highly overlap with the these of Faster-RCNN. Instead, we drew the performance of SSD in Figure \ref{fig_Reso} which has more space. Table \ref{tbl_compare1} shows the areas under curve of the ROC curves. Our method is also substantially better than the second best Faster-RCNN method. 

We also analyze the performance on players with different pixel resolutions in Figure \ref{fig_Reso}. Our method is more robust than other methods when pixel resolutions decline. If we compare Figure \ref{fig_Reso} with Figure \ref{fig:wihout_end_to_end}, we can find the performance gain in small players are from the end-to-end learning and the dilation strategy.

\begin{figure}[htpb]
	\centering
	\includegraphics[width=1.0\linewidth]{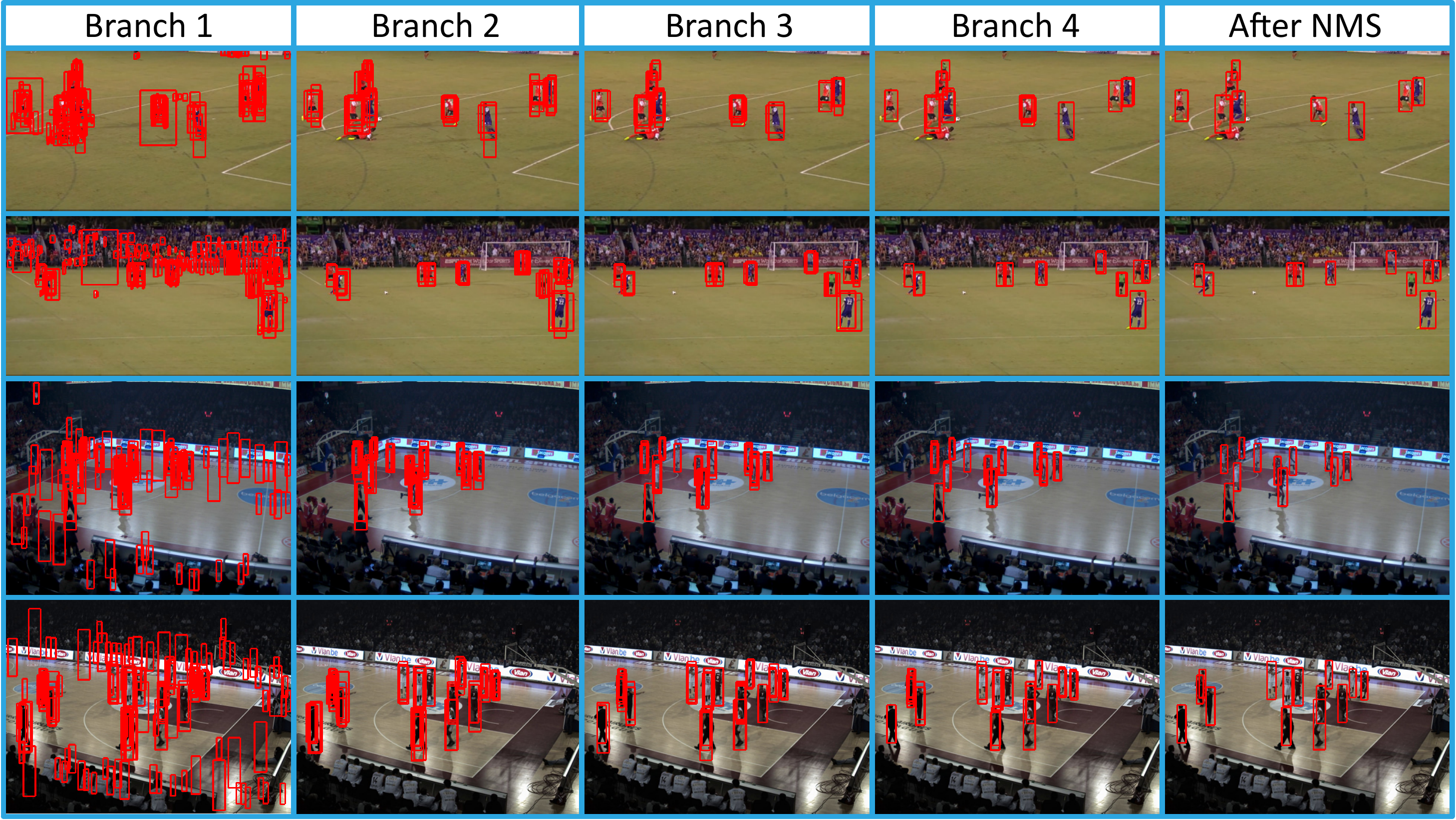}	    
	\vspace{-4ex}
	\caption{\textbf{Qualitative Results.} The proposed method is able to effectively reject non-player samples stage by stage and output accurate bounding boxes for players in both soccer and basketball games. In this figure, NMS stands for non-maximum suppression.}
	\label{fig_Qualitative}
\end{figure}

{Figure \ref{fig_Qualitative} shows some qualitative results of our method. Our method performs robustly among various conditions. However, it may can not produce accurate bounding-boxes when players have complicated postures (e.g., the laying down players). One way to solve this problem is to integrate bounding-boxes regression strategy into our cascaded CNN, which will be an interesting direction for future work.}

%\vspace{-2mm}
\subsection{Implementation}
We perform experiments on a laptop with an Intel i7-6700HQ (2.6GHz) processor and a NVIDIA GTX1060 GPU. The detection speed is about 10 fps for images of size 1280\(\times\)720 using un-optimized Matlab code based on MatConvNet~\cite{vedaldi15matconvnet}. Because our architecture achieves state-of-the-art performance while being much lighter (i.e., it has much fewer parameters) compared with conventional CNNs, it has the potential to be more efficient by optimizing the implementation.

%\vspace{-2mm}
\section{Conclusions and future work}
\label{sec:conclusion}

We have presented an accurate and efficient approach for player detection in group sports. We first introduced a light and effective cascaded CNN where all cascade stages are globally optimized end-to-end. Then we presented a dilation strategy to improve the detection accuracy of the network when performed on a whole image. In addition, we proposed a soccer player dataset to evaluate the robustness of player detection algorithms. Experimental results on both soccer and basketball datasets suggest that the proposed approach is light, effective and robust compared with many state-of-the-art detection methods. 

Player detection has not been fully solved in terms of localization accuracy. In the future, we would like to employ regression strategies to obtain more accurate bounding boxes for players that have complicated postures. Our method is designed for applications with relative-simple backgrounds, which generally holds for sports applications. We have not test the method on varied backgrounds datasets such as Caltech pedestrian dataset, which we leave as a future work.

\bibliography{egbib}

\end{document}